\documentclass[10pt,twocolumn,letterpaper]{article}

\usepackage{cvpr}
\usepackage{times}
\usepackage{subfig}
\usepackage{graphicx}
\usepackage{amsmath}
\usepackage{amssymb}
\usepackage{multirow}
\usepackage{epsfig}
\usepackage{gensymb}
\usepackage{tabularx, booktabs}

\newcolumntype{Y}{>{\arraybackslash}X}

\usepackage{pifont}
\usepackage{caption}
\usepackage[boxed,algoruled,linesnumbered]{algorithm2e}
\def\mbf#1{\mathbf{#1}}

\def\tbf#1{\textbf{#1}}

\newcommand{\minisection}[1]{\vspace{1mm}\noindent{\bf #1}}
\newcommand{\cmark}{\ding{51}}%
\newcommand{\xmark}{\ding{53}}%

\usepackage[pagebackref=true,breaklinks=true,letterpaper=true,colorlinks,bookmarks=false]{hyperref}

\cvprfinalcopy 

\ifcvprfinal\fi
\begin{document}

\title{Regressing Robust and Discriminative 3D Morphable Models\\with a very Deep Neural Network\vspace{-3mm}}

\author{Anh Tu\~{a}n Tr\~{a}n$^1$, Tal Hassner$^{2,3}$, Iacopo Masi$^1$, and G\'{e}rard Medioni$^1$\\
{\small $^{1}$ Institute for Robotics and Intelligent Systems, USC, CA, USA}\\
{\small $^{2}$ Information Sciences Institute, USC, CA, USA}\\
{\small $^{3}$ The Open University of Israel, Israel}}

\maketitle

\begin{abstract}
The 3D shapes of faces are well known to be discriminative. Yet despite this, they are rarely used for face recognition and always under controlled viewing conditions. We claim that this is a symptom of a serious but often overlooked problem with existing methods for single view 3D face reconstruction: when applied ``in the wild'', their 3D estimates are either unstable and change for different photos of the same subject or they are over-regularized and generic. In response, we describe a robust method for regressing discriminative 3D morphable face models (3DMM). We use a convolutional neural network (CNN) to regress 3DMM shape and texture parameters directly from an input photo. We overcome the shortage of training data required for this purpose by offering a method for generating huge numbers of labeled  examples. The 3D estimates produced by our CNN surpass state of the art accuracy on the MICC data set. Coupled with a 3D-3D face matching pipeline, we show the first competitive face recognition results on the LFW, YTF and IJB-A benchmarks using 3D face shapes as representations, rather than the opaque deep feature vectors used by other modern systems.\vspace{-5mm}
\end{abstract}

\section{Introduction}
Single view 3D face shape estimation methods originally proposed using their 3D shapes for recognition~\cite{blanz2002face,blanz2003face,paysan09basel}. This makes sense because 3D shapes are discriminative -- different people have different face shapes -- yet invariant to lighting, texture changes and more. Indeed, previous work showed that when available, high resolution 3D face scans are excellent face representations which can even be used to distinguish between the faces of identical twins~\cite{bronstein2005three}. 

Curiously, however, despite their widespread use, single view face reconstruction methods are rarely employed by modern face recognition systems. The highly successful 3D Morphable Models (3DMM), for example, were only ever used for recognition in limited, controlled viewing conditions~\cite{blanz2002face,blanz2003face,chu2014,hu2016face,paysan09basel}. To our knowledge, there are no reports of successfully using single view face shape estimation -- 3DMM or any other method -- to recognize faces in challenging unconstrained, {\em in the wild} settings.

\begin{figure}[t]
\centering
\includegraphics[width=\columnwidth]{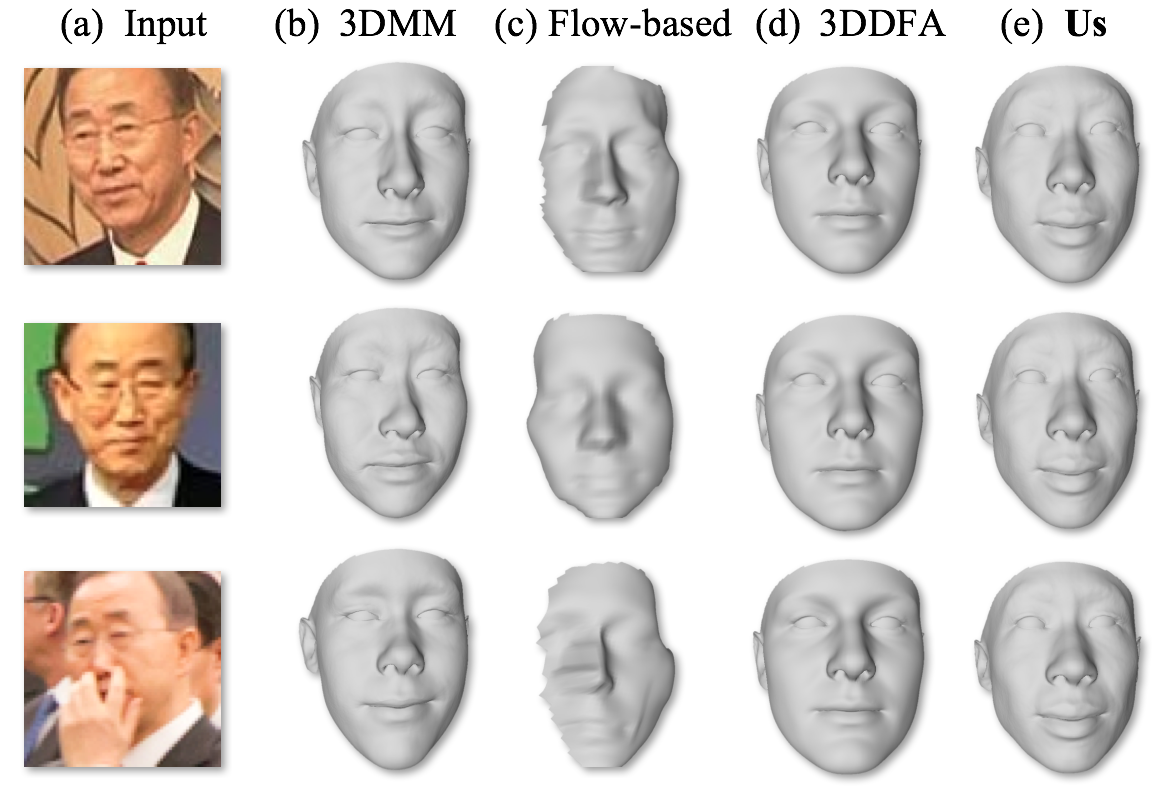}\vspace{-2mm}
\caption{{\em Unconstrained, single view, 3D face shape reconstruction}. (a) Input images of the same subject with disruptive poses and occlusions. (b-e) 3D reconstructions using (b) single-view 3DMM~\cite{romdhani2005estimating}, (c) flow based method~\cite{hassner2013viewing} (d) 3DDFA~\cite{Zhu2016Face}, (e) Our proposed approach. (b-c) Present different 3D shapes for the same subject and (d) appears generic, whereas our method (e) is robust, producing similar discriminative 3D shapes for different views.}
\label{fig:teaser}
\vspace{-4mm}
\end{figure}

An important reason why this maybe so, is that these methods can be unstable in unconstrained viewing conditions. We later verify this quantitatively but it can also be seen in Fig.~\ref{fig:teaser} which presents 3D shapes estimated from three unconstrained photos by three different methods (Fig.~\ref{fig:teaser}~(b-d)). Clearly, though the same subject appears in all photos, {\em shapes produced by the same method are either very different (b,c) or highly regularized and generic (d)}. It is therefore unsurprising that these shapes are poor representations for recognition. It also explains why some recently proposed using coarse, simple 3D shape approximations only as proxies when rendering faces to new views rather than as face representations~\cite{hassner2013viewing,hassner2015effective,masi16dowe,taigman2014deepface}.

Contrary to previous work, we show that robust and discriminative 3D face shapes can, in fact, be estimated from single, unconstrained images (Fig.~\ref{fig:teaser}~(e)). We propose estimating 3D facial shapes using a very deep convolutional neural network (CNN) to regress 3DMM shape and texture parameters directly from single face photos. We identify shortage of labeled training data as an obstacle to using data-hungry CNNs for this purpose. We address this problem with a novel means for generating a huge labeled training set of unconstrained faces and their 3DMM representations. Coupled with additional technical novelties, we obtain a method which is fast, robust and accurate.

The accuracy of our estimated shapes is verified on the MICC data set~\cite{bagdanov11micc} and {\em quantitatively shown to surpass the accuracy of other 3D reconstruction methods}. We further show that our estimated shapes are robust and discriminative by presenting face recognition results on the Labeled Faces in the Wild (LFW)~\cite{LFWTech}, YouTube Faces (YTF)~\cite{wolf:YTF} and IJB-A~\cite{Klare_2015_CVPR} benchmarks. To our knowledge, {\em this is the first time single image 3D face shapes are successfully used to represent faces from modern, unconstrained face recognition benchmarks}. Finally, to promote reproduction of our results, we publicly release our code and models.\footnote{Please see~\url{www.openu.ac.il/home/hassner/projects/CNN3DMM} for updates.}.

\section{Related work}\label{sec:related}
Over the years, many attempts were made to estimate the 3D surface of a face appearing in a single view. Before listing them, it is important to mention recent {\em multi image} methods which use image sets for reconstruction (e.g.,~\cite{liang2016head,Piotraschke_2016_CVPR,Roth_2015_CVPR,Roth_2016_CVPR,suwajanakorn2014total}). Although these methods produce accurate 3D reconstructions, they require many images from multiple sources to produce a single 3D face shape whereas we reconstruct faces from single images.

Methods for {\em single view} 3D face reconstructions can broadly be categorized into the following types. 

\minisection{Statistical shape representations}, such as the widely popular 3DMM~\cite{blanz2004exchanging,blanz1999morphable,chu2014,paysan09basel,romdhani2003efficient,tang2008real,yang2011expression}, use many aligned 3D face shapes to learn a  distribution of 3D faces, represented as a high dimensional subspace. Each point on this subspace is a parameter vector representing facial geometry and sometimes expression and texture. Reconstruction is performed by searching for a point on this subspace that represents a face similar to the one in the input image. These methods do not attempt to produce discriminative facial geometries and indeed, as mentioned earlier, were only used for face recognition under controlled settings.

The very recent method of~\cite{richardson20163d} also uses a CNN to regress 3DMM parameters for face photos. They too recognize absence of training data as a major concern. Contrary to us, they propose synthesizing training faces with known geometry by sampling from the 3DMM distribution. This approach produces synthetic looking photos which can easily cause overfitting problems when training large networks~\cite{masi16dowe}. They were therefore able to train only a shallow residual network (seven layers compared to our 101) and their estimated shapes were not shown to be more robust or discriminative than other methods.

\minisection{Scene assumption methods.} In order to obtain {\em correct} reconstructions, some make strong assumptions on the scene and the viewing conditions in the input image. Shape from shading methods~\cite{kemelmacher20113d}, for example, make assumptions on the light sources, facial reflectance and more. Others instead use facial symmetry~\cite{dovgard2004statistical}. The assumptions they and others make often do not hold in practice, limiting the application of these methods to controlled settings. 

\minisection{Example based methods}, beginning from the work of~\cite{hassner2006example} and more recently~\cite{hassner2013viewing,taigman2014deepface}, modify the 3D surface of example face shapes, fitting them to the face appearing in input photo. These methods favor robustness to challenging viewing conditions over detailed reconstructions. They were thus only used for face recognition to synthesize new views from unseen poses.

\minisection{Landmark fitting methods.} Finally, some reconstruction techniques fit a 3D surface to detected facial landmarks rather than to face intensities directly. These include methods designed for videos (e.g.,~\cite{huber:3dmm,saito2016}) and the CNN based approaches of~\cite{jourabloo2016large,Zhu2016Face}. These focus more on landmark detection than 3D shape estimation and so do not attempt to produce detailed and discriminative facial geometries. 

\section{Regressing 3DMM parameters with a CNN}
We propose to regress 3DMM face shape parameters directly from an input photo using a very deep CNN. Ostensibly, CNNs are ideal for this task: After all, they are being successfully applied to many related computer vision tasks. But despite their success, apart from~\cite{richardson20163d}, we are unaware of published reports of using CNNs for 3DMM parameter regression.

We believe CNNs were not used here because this is a regression problem where both the input photo and the output 3DMM shape parameters are high dimensional. Solving such problems requires deep networks and these need massive amounts of training data. Unfortunately, existing unconstrained face sets with ground truth 3D shapes are far too small for this purpose and obtaining large quantities of 3D face scans is labor intensive and impractical. 

We therefore instead leverage three key observations. 
\begin{enumerate}
\item As discussed in Sec.~\ref{sec:related}, accurate 3D estimates can be obtained by using {\em multiple images} of the same face.
\item Unlike the limited availability of ground truth 3D face shapes, there is certainly no shortage of challenging face sets containing multiple photos per subject.
\item Highly effective deep networks are available for the related task of extracting robust and discriminative face representations for face recognition. 
\end{enumerate}
From (1), we have a reasonable way of producing 3D face shape estimates for training, as surrogates for ground truth shapes: by using a robust method for multi-view 3DMM estimation. Getting multiple photos for enough subjects is very easy (2). This abundance of examples further allows balancing any reconstruction errors with potentially limitless subjects to train on. Finally, (3), a state of the art CNN for face recognition may be fine-tuned to this problem. It should already be tuned for unconstrained facial appearance variations and trained to produce similar, discriminative outputs for different images of the same face. 

\begin{figure*}[t]
\centering

\includegraphics[width=.93\textwidth]{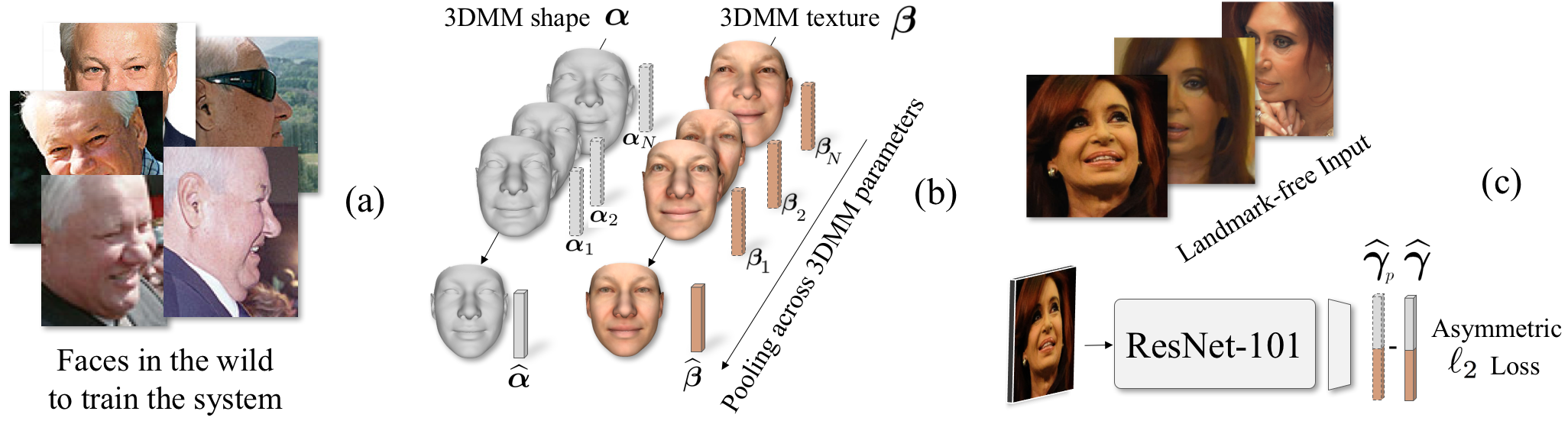}\vspace{-2mm}
\caption{{\em Overview of our process.} (a) Large quantities of unconstrained photos are used to fit a single 3DMM for each subject. (b) This is done by first fitting single image 3DMM shape and texture parameters to each image separately. Then, all 3DMM estimates for the same subject are pooled together for a single estimate per subject. (c) These pooled estimates are used in place of expensive ground truth face scans to train a very deep CNN to regress 3DMM parameters directly. }
\label{fig:system}
\vspace{-4mm}
\end{figure*}

\subsection{Generating training data}\label{sec:gendata}
To generate training data, we use a simple yet effective {\em multi image} 3DMM estimation method, loosely based on the one recently proposed by~\cite{Piotraschke_2016_CVPR}. We run it on the unconstrained faces in the CASIA WebFace dataset~\cite{yi2014learning}. These multi image 3DMM estimates are then used as ground truth 3D face shapes when training our CNN 3DMM regressor. 

Multi image 3DMM reconstruction is performed by first estimating 3DMM parameters from the 500k single images in CASIA. 3DMM estimates for images of the same subject are then aggregated into a single 3DMM per subject ($\sim$10k subjects). This process is described next (see also, Fig.~\ref{fig:system}).

\minisection{The 3DMM representation.} 
Our system uses the popular Basel Face Model (BFM)~\cite{paysan09basel}. It is a publicly available 3DMM representation and one of the state of the art methods for single view 3D face modeling. 

A face is modeled by decoupling its shape and texture giving the following two independent generative models.
\begin{equation}
\mbf{S}^{\prime} = \widehat{\mbf{s}} + \mbf{W}_S~\boldsymbol{\alpha}  \quad,  \quad
\mbf{T}^{\prime} = \widehat{\mbf{t}} + \mbf{W}_T~\boldsymbol{\beta}.
\label{eq:3DMM}
\end{equation}
Here, the vectors $\widehat{\mbf{s}}$ and $\widehat{\mbf{t}}$ are the mean face shape and texture, computed over the aligned facial 3D scans in the Basel Faces collection and represented by the concatenated 3D coordinates of the 3D point clouds and the concatenated RGB values of their textures. Matrices $\mbf{W}_S$ and $\mbf{W}_T$ are the principle components, computed from the same aligned facial scans. Finally, $\boldsymbol{\alpha}$ and $\boldsymbol{\beta}$ are each 99D parameter vectors, representing shape and texture respectively. 

\minisection{Single image 3DMM fitting.} Fitting a 3DMM to each training  image is performed with a slightly modified version of the two standard methods of~\cite{blanz:tpami:03} and~\cite{romdhani2005estimating}. Given an image $\mbf{I}$, we estimate parameter vectors $\boldsymbol{\alpha}^\star$ and $\boldsymbol{\beta}^\star$ which represent a face similar to the one in~$\mbf{I}$ (Eq.~(\ref{eq:3DMM})). Unlike previous work, we begin processing by applying the CLNF~\cite{Kanggeon:bmvc16} state of the art facial landmark detector. It provides $K=68$ facial landmarks $\mbf{p}_k \in \mathbb{R}^{2}$, $k\in 1..K$, and a confidence score value $w$ (which we use later on).

Landmarks are used to obtain an initial estimate for the pose of the input face, in the reference 3DMM coordinate system. Pose is represented by six degrees of freedom for rotation, $\mbf{r}=[r_\alpha, r_\beta, r_\gamma]$, and translation, $\mbf{t}=[t_X, t_Y, t_Z]$, and estimated similar to~\cite{hassner2013viewing}. 3DMM fitting then proceeds by optimizing over the shape, texture, pose, illumination, and color model following~\cite{blanz:tpami:03}. We found that CLNF makes occasional localization errors. To introduce more stability, our optimization also uses the edge-based cost of~\cite{romdhani2005estimating}. For more details on this optimization, we refer to~\cite{blanz:tpami:03} and~\cite{romdhani2005estimating}. 

Once the optimization converges, we take the shape and texture parameters, $\boldsymbol{\alpha}^\star$ and $\boldsymbol{\beta}^\star$, from the last iteration as our single image 3DMM estimate for the input image $\mbf{I}$. Importantly, though this process is known to be computationally expensive, it is applied in our pipeline only in preprocessing and once for every training image. We later show our CNN regressor to be much faster. 

\minisection{Multi image 3DMM fitting.} Although a number of multi image 3D face shape estimation methods were proposed in the past, we found the following simple approach, inspired by the very recent work of~\cite{Piotraschke_2016_CVPR}, to be particularly effective. 

Specifically, we pool the shape and texture 3DMM parameters $\boldsymbol{\gamma}_i=[\boldsymbol{\alpha}_i,\boldsymbol{\beta}_i], i\in 1..N$ across all the $N$ single view estimates belonging to the same subject. Pooling is performed by element wise weighted averaging of the $N$ 3DMM vectors, resulting in a single 3DMM estimate for that subject, $\widehat{\boldsymbol{\gamma}}$. That is,
\begin{equation}
\widehat{\boldsymbol{\gamma}} = \sum_{i=1}^N w_i\cdot\boldsymbol{\gamma}_i \quad \textnormal{and} \quad \sum_{i=1}^N w_i=1,
\label{eq:pooling}
\end{equation}
where $w_i$ are normalized per-image confidences provided by the CLNF facial landmark detector. 

Note that unlike~\cite{Piotraschke_2016_CVPR}, we do not use a rank-list based on distances of normals as a quality measure to pool 3DMM parameters, instead taking the landmark detection confidence measure for these weights. Following this process, each CASIA subject is associated with a single, pooled 3DMM parameter vector $\widehat{\boldsymbol{\gamma}}$. For ease of notation, henceforth we will drop the {\em hat} when denoting pooled features, assuming all training set 3DMM parameters were pooled. 

\subsection{Learning to regress pooled 3DMM} \label{sec:learning}
Following the process described in Sec.~\ref{sec:gendata}, each subject in our data set is associated with a number of images and a single, pooled 3DMM. We now use this data to learn a function which, ideally, regresses the \emph{same} pooled 3DMM feature vector for different photos of the same subject. 

To this end, we use a state of the art CNN, trained for face recognition. We use the very deep ResNet architecture~\cite{He_2016_CVPR} with 101 layers, recently trained for face recognition by~\cite{masi16dowe}. We modify its last fully-connected layer to output the 198D 3DMM feature vector $\boldsymbol{\gamma}$. The network is then fine-tuned on CASIA images using the pooled 3DMM estimates as target values; different images of the same subject presented to the CNN using the same target 3DMM shape. We note that we also tried using the VGG-Face CNN of~\cite{parkhi2015deep} with 16 layers. Its results were similar to those obtained by the ResNet architecture, though somewhat lower. 

\begin{figure}[t]
\centering
\includegraphics[width=\columnwidth]{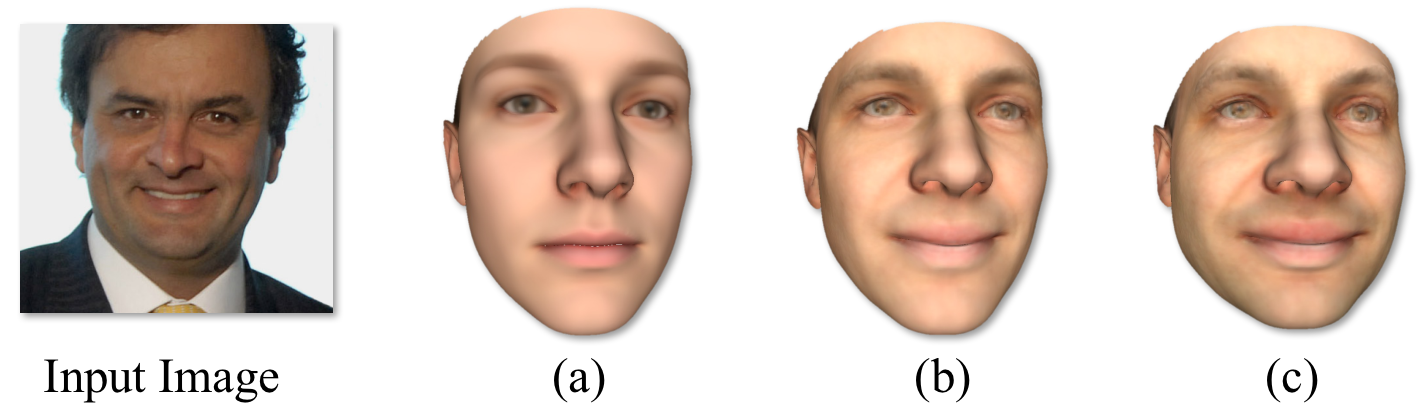}
\caption{{\em Effect of our loss function:} (left) Input image, (a) generic model, (b) regressed shape and texture with a regular $\ell_2$ loss and (c) our proposed asymmetric $\ell_2$ loss.}
\label{fig:qual_loss}
\vspace{-4mm}
\end{figure}

\minisection{The asymmetric Euclidean loss.} Training our network requires some care when defining its loss function. 3DMM vectors, by construction, belong to a multivariate Gaussian distribution with its mean on the origin, representing the mean face (Sec.~\ref{sec:gendata}). Consequently, during training, using the standard Euclidean loss to minimize distances between estimated and target 3DMM vectors will favor estimates closer to the origin: these will have a higher probability of being closer to their target values than those further away. In practice, we found that a network trained with the Euclidean loss tends to output less detailed faces (Fig.~\ref{fig:qual_loss}).

To counter this bias towards a mean face shape, we introduce an {\em asymmetric Euclidean loss}. It is designed to encourage the network to favor estimates further away from the origin by decoupling under-estimation errors (errors on the side of the 3DMM target closer to the origin) from over-estimation errors (where the estimate is further out from the origin than the target). It is defined by:
\vspace{-3mm}

\begin{equation}
\mathcal{L}(\boldsymbol{\gamma}_p,\boldsymbol{\gamma}) =  \lambda_1 \cdot  \underbrace{\vert\vert \boldsymbol{\gamma}^+ - \boldsymbol{\gamma}_{\textnormal{max}}\vert\vert_2^2}_{\text{over-estimate}}+ \lambda_2 \cdot  \underbrace{\vert\vert \boldsymbol{\gamma}^+_p - \boldsymbol{\gamma}_{\textnormal{max}}\vert\vert_2^2}_{\text{under-estimate}}, 
\label{eq:loss}
\end{equation}
using the element-wise operators:
\begin{align}\vspace{-2mm}
\boldsymbol{\gamma}^+\doteq \textnormal{abs}(\boldsymbol{\gamma}) \doteq \textnormal{sign}(\boldsymbol{\gamma}) \cdot \boldsymbol{\gamma};\quad \boldsymbol{\gamma}_p^+\doteq  \textnormal{sign}(\boldsymbol{\gamma}) \cdot \boldsymbol{\gamma}_p,\\
\boldsymbol{\gamma}_{\textnormal{max}} \doteq \max(\boldsymbol{\gamma}^+,\boldsymbol{\gamma}^+_p).
\end{align}
Here, $\boldsymbol{\gamma}$ is the target pooled 3DMM value, $\boldsymbol{\gamma}_p$ is the output, regressed 3DMM and $\lambda_{1,2}$ control the trade-off between the over and under estimation errors. When both equal 1, this reduces to the traditional Euclidean loss. In practice, we set $\lambda_{1}=1$, $\lambda_{2}=3$, thus changing the behavior of the training process, allowing it to escape under-fitting faster and encouraging the network to produce more detailed, realistic 3D face models (Fig.~\ref{fig:qual_loss}).

\minisection{Network hyperparameters.} Eq.~(\ref{eq:loss}) is solved using Stochastic Gradient Descent (SGD) with a mini-batch of size 144, momentum set to 0.9 and with regularization over the weights provided by $\ell_2$ with a weight decay of 0.0005. When performing back-propagation, we learn the inner product layer (fc) after pool5 faster, setting the learning rate to $0.01$, since it is trained from scratch for the regression problem. Other network weights are updated with a learning rate an order of magnitude lower. When the validation loss saturates, we decrease learning rates by an order of magnitude, until the validation loss stops decreasing.

\minisection{Discussion: Render-free 3DMM estimator.} It is important to note that by choosing to use a CNN to regress 3DMM parameters, we obtain a function that is {\em render-free}. That is, 3DMM parameters are regressed directly from the input image, without an optimization process which renders the face and compares it to the photo, as do existing methods for 3DMM estimation (including our method for generating training data in Sec.~\ref{sec:gendata}). By using a CNN, we therefore hope to gain not only improved accuracy, but also much faster 3DMM estimation speeds.

\subsection{Parameter based 3D-3D recognition} \label{sec:matching}
The CNN we train in Sec.~\ref{sec:learning} represents a function $f:\mbf{I} \mapsto \boldsymbol{\gamma}_p$, giving us 3DMM parameters $\boldsymbol{\gamma}_p$ for an input image $\mbf{I}$. We later use our 3DMM estimates in face recognition benchmarks, to test how robust and discriminative they are. We next describe the method used for that purpose to evaluate the similarity of two face shapes and textures to determine if they represent the same subject.

\minisection{3D-3D recognition with a single image.} We perform face recognition using the 3DMM parameters regressed by our network: By using the 3DMM parameters $\boldsymbol{\gamma}_p$ as face descriptors. Because different benchmarks often exhibit specific appearance biases, we apply Principal Component Analysis (PCA), learned from the training splits of the test benchmark, to adapt our estimated parameter vectors to the benchmark. Signed, element wise square rooting of these vectors is then used to further improve representation power~\cite{perronnin2010improving}. Finally, the similarity of two faces, $s(\boldsymbol{\gamma}_{p1},\boldsymbol{\gamma}_{p2})$, is evaluated by computing their cosine score:\vspace{-1mm}

\begin{equation}
s(\boldsymbol{\gamma}_1,\boldsymbol{\gamma}_2) = \frac{\boldsymbol{\gamma}_{p1} \cdot \boldsymbol{\gamma}_{p2}^T}{\vert\vert
\boldsymbol{\gamma}_{p1}\vert\vert \cdot \vert\vert\boldsymbol{\gamma}_{p2}\vert\vert}.
\label{eq:sim}
\end{equation}
\minisection{3D-3D recognition with multiple-images.} In some scenarios, a subject is represented by a set of images, rather than just one. This is the case in the YTF benchmark~\cite{wolf:YTF} where videos are used, each containing multiple frames, and in the recent IJB-A~\cite{Klare_2015_CVPR}, which uses {\em templates} containing heterogeneous visual data (images, videos and possibly more). 

We use the same pipeline for single images also for image sets. Here, however, 3DMM parameters for different images or frames are first pooled using Eq.~(\ref{eq:pooling}). Unlike the process applied in Sec.~\ref{sec:gendata}, all images here have equal weights, as we do not run landmark detection prior to 3DMM fitting with our CNN (see below). When using templates with both videos and images, following~\cite{masi16dowe}, we first pool the 3DMM estimates for frames in each video separately, obtaining one 3DMM per video. We then pool these 3DMMs with those of other images in the same template. 

\minisection{Face alignment.} Facial landmark detection and face alignment are known to improve recognition accuracy (e.g.,~\cite{WHT:ECCVW08:DBMW,hassner2015effective}). In fact, the recent, related work of~\cite{hu2016face} manually assigned landmarks before using their 3DMM fitting method for recognition on controlled images. We, however, {\em did not align faces} beyond using the bounding boxes provided in their data sets. We found our method robust to misalignments and so spared the runtime this required. 
\vspace{-2mm}
\section{Experimental results}\label{sec:results}
We test our proposed method, comparing the accuracy of its estimated 3D shapes, its speed and its ability to represent faces for recognition with existing methods. Importantly, {\em we are unaware of any previous work on single view 3D face shape estimation which reported as many quantitative tests as we do}, in terms of the number of benchmarks used, the number of baseline methods compared with and the level of difficulty of the photos used in these tests. 

Specifically, we evaluate the accuracy of our estimated 3D shapes using videos and photos and their corresponding scanned, ground truth 3D shapes from the MICC Florence Faces dataset~\cite{bagdanov11micc} (Sec.~\ref{sec:exp:acc}). To test how discriminative and robust our shapes are when estimated from unconstrained images, we perform single image and multi image face recognition using the LFW~\cite{LFWTech}, YTF~\cite{wolf:YTF} and the new IARPA JANUS Benchmark-A (IJB-A)~\cite{Klare_2015_CVPR} (Sec.~\ref{sec:exp:rec}). Finally we also provide qualitative results in Sec.~\ref{sec:exp:qual}.

As baseline 3D reconstruction methods we used standard 3DMM fitting~\cite{romdhani2005estimating}, which we implemented ourselves, the flow-based method of~\cite{hassner2013viewing}, the edge based method of~\cite{Bas:accvw16}, the multi resolution, multi-view approach of~\cite{huber:3dmm} and the recent 3DDFA~\cite{Zhu2016Face}, were all tested with their authors' implementations.

\subsection{3D shape reconstruction accuracy}\label{sec:exp:acc} The MICC dataset~\cite{bagdanov11micc} contains challenging face videos of 53 subjects. The videos span the range of controlled to challenging unconstrained outdoor settings. For each of the subjects in these videos, the data set contains also a ground-truth 3D model acquired using a structured-light scanning system with high precision. This allows comparing our 3D face shape estimates with the ground truth shapes. 
\begin{table}[t]
\centering
\resizebox{1.0\linewidth}{!}{
\setlength{\tabcolsep}{1pt}
\begin{tabular}{l ccccc}
\toprule
Method & 3DRMSE  &	RMSE &	$\log_{10}\times 10^4$ & Rel$\times 10^4$ & Sec. \\
\cmidrule(r){1-1} \cmidrule(l){2-6}
Generic & 1.88$\pm$.52 & 3.48$\pm$.76 & 28$\pm$7 & 65$\pm$16 & -- \\
3DMM~\cite{romdhani2005estimating} & 1.75$\pm$.42 & 3.64$\pm$.94 &29$\pm$8 &68$\pm$18  & 120\\
Flow-based~\cite{hassner2013viewing} & 1.83$\pm$.39 & 3.29$\pm$.70 & 27$\pm$6 & 62$\pm$14 & 13.3 \\
\cmidrule(r){1-1} 
Us & \tbf{1.57$\pm$.33} & \tbf{3.18$\pm$.77} & \tbf{26$\pm$6} & \tbf{59$\pm$14}  & \tbf{.088} \\
\cmidrule(r){1-1} \cmidrule(l){2-6}
Generic\emph{+pool} & 1.88$\pm$.52 & 3.48$\pm$.76 & 28$\pm$7 & 65$\pm$16 & -- \\
3DMM~\cite{romdhani2005estimating}\emph{+pool}$^*$ & 1.60$\pm$.46 & 3.31$\pm$.98 & 27$\pm$9 & 62$\pm$20 & 120 \\
3DDFA~\cite{Zhu2016Face}\emph{+pool}& 1.83$\pm$.58 & 3.45$\pm$.85 & 28$\pm$7 & 65$\pm$17 & .146 \\
\cite{huber:3dmm} & 1.84$\pm$.32 & 3.73$\pm$.62 & 30$\pm$5 & 68$\pm$11 & .372 \\
\cite{Bas:accvw16}\emph{+pool} & 1.84$\pm$.58 & 3.45$\pm$.85 & 28$\pm$6 & 65$\pm$13 & 52.3 \\

\cmidrule(r){1-1}
Us \emph{+pool} &\tbf{1.53$\pm$.29}&\tbf{3.14$\pm$.70}& \tbf{25$\pm$6} & \tbf{58$\pm$13} & \tbf{.088}\\
\bottomrule
\end{tabular}
}
\caption{{\em 3D estimation accuracy and per-image speed} on the MICC dataset. Top are single view methods, bottom are multi frame. See text for details on measures. 3DRMSE in real-world $mm$; $\log_{10}$ and Rel were both scaled to preserve space. $^*$ Denotes the method used to produce the training data in Sec.~\ref{sec:gendata}. Lower values are better.}
\vspace{-2mm}
\label{tab:micc}
\end{table}
 
\begin{table*}[tb]
\centering
\resizebox{.94\linewidth}{!}{
\begin{tabular}{lccccccc}
\toprule
Method &	3D & Texture	& Accuracy	& 100\%-EER &	AUC	 & TAR-10\%	& TAR-1\% \\ \hline
\multicolumn{7}{c}{\tbf{Labeled Faces in the Wild}} \\
EigenFaces~\cite{turk:eigface} & \multicolumn{2}{c}{--} & 60.02$\pm$0.79 & -- & -- & 25 & 6.2 \\
Hybrid Descriptor~\cite{WHT:ECCVW08:DBMW} & \multicolumn{2}{c}{--} & 78.47$\pm$0.51 & -- & -- & 66.60 & 42.4 \\
DeepFace-ensemble~\cite{taigman2014deepface} & \multicolumn{2}{c}{--} & 97.35$\pm$0.25 & -- & -- & 99.6 & 93.7 \\ 
AugNet~\cite{masi16dowe} & \multicolumn{2}{c}{--} & 98.06$\pm$0.60 & 98.00$\pm$0.73 & -- & 99.5 & 94.2 \\ 
\cmidrule(r){1-1} \cmidrule(l){2-8}
\multirow{3}{*}{3DMM~\cite{romdhani2005estimating}} & \cmark & \xmark &  66.13$\pm$2.79 & 65.70$\pm$2.81 & 72.24$\pm$2.75 & 35.90$\pm$3.74 & 12.37$\pm$4.81 \\
 & \xmark & \cmark & 74.93$\pm$1.14 & 74.50$\pm$1.21 & 82.94$\pm$1.14 & 60.40$\pm$3.15 & 28.73$\pm$7.17 \\
 & \cmark & \cmark & 75.25$\pm$2.12 & 74.73$\pm$2.56 & 83.21$\pm$1.93 & 59.4$\pm$4.64 & 29.67$\pm$4.73 \\
\cmidrule(r){1-1}
3DDFA~\cite{Zhu2016Face}  & \cmark & \xmark &  66.98$\pm$2.56 & 67.13$\pm$1.90 & 73.30$\pm$2.49 & 36.76$\pm$6.27 & 10.00$\pm$3.22 \\
\cmidrule(r){1-1}
\multirow{3}{*}{Us} & \cmark & \xmark &  90.53$\pm$1.34 & 90.63$\pm$1.61 & 96.6$\pm$0.79 & 91.13$\pm$2.62 & 58.20$\pm$12.14 \\
 & \xmark & \cmark & 90.6$\pm$1.07 & 90.70$\pm$1.17 & 96.75$\pm$0.59 & 91.23$\pm$2.42 & 52.60$\pm$8.14 \\
 & \cmark & \cmark & \tbf{92.35$\pm$1.29} & \tbf{92.33$\pm$1.33} & \tbf{97.71$\pm$0.64} & \tbf{94.2$\pm$2.00} & \tbf{65.57$\pm$6.93} \\
\bottomrule
\multicolumn{7}{c}{\tbf{YouTube Faces}} \\
MBGS LBP~\cite{wolf:YTF} & \multicolumn{2}{c}{--} & 76.4$\pm$1.8 & 74.7 &  82.6 & 60.5 & 35.8 \\
DeepFace-ensemble~\cite{taigman2014deepface} & \multicolumn{2}{c}{--} & 91.4$\pm$1.1 & 91.4 & 96.3  & 92 & 54 \\ 
\cmidrule(r){1-1} \cmidrule(l){2-8}
\multirow{3}{*}{3DMM~\cite{romdhani2005estimating}+{\em pool}$^*$} & \cmark & \xmark & 73.26$\pm$2.51 & 73.08$\pm$2.65 & 80.41$\pm$2.60 & 51.36$\pm$5.11 & 24.04$\pm$4.56 \\
& \xmark & \cmark & 77.34$\pm$2.54 & 76.96$\pm$2.64 & 85.32$\pm$2.63 & 63.16$\pm$5.07 & 31.36$\pm$5.21 \\
& \cmark & \cmark & 79.56$\pm$2.08 & 79.20$\pm$2.07 & 87.35$\pm$1.92 & 69.08$\pm$5.00 & 34.56$\pm$6.89 \\
\cmidrule(r){1-1}
3DDFA~\cite{Zhu2016Face}+{\em pool}& \cmark & \xmark & 68.10$\pm$2.93 & 67.96$\pm$3.12 & 74.95$\pm$3.04 & 40.52$\pm$3.65 & 12.2$\pm$2.67 \\
\cmidrule(r){1-1}
\multirow{3}{*}{Us \emph{+pool}}
& \cmark & \xmark &  88.28$\pm$1.84 & 88.32$\pm$2.16 & \tbf{95.95$\pm$1.38} & 86.60$\pm$3.95 & \tbf{51.12$\pm$8.86} \\
& \xmark & \cmark  & 87.56$\pm$2.56 & 87.68$\pm$2.25 & 94.44$\pm$1.38 & 84.80$\pm$4.89 & 40.92$\pm$8.26 \\ 
& \cmark & \cmark & \tbf{88.80$\pm$2.21} & \tbf{88.84$\pm$2.40}  & 95.37$\pm$1.43 & \tbf{87.92$\pm$4.18} & 46.56$\pm$6.20 \\ 
\bottomrule
\end{tabular}
}
\caption{{\em LFW and YTF face verification.} Comparing our 3DMM regression with others, including baseline face recognition methods. $^*$ Denotes the same method used to produce 3DMM target values for our CNN training (Sec.~\ref{sec:gendata}).}
\vspace{-3mm}
\label{tab:lfw-ytf}
\end{table*}

\begin{table}[tb]
\centering
\footnotesize
\setlength{\tabcolsep}{.2pt}
\begin{tabularx}{\columnwidth}{lccl*{4}{Y}}
\toprule
Method &	3D	& Text. & TAR-10\%	& TAR-1\% &	Rank-1 & Rank-5 & Rank-10 \\
\cmidrule(r){1-1} \cmidrule(l){2-8}
AugNet & \multicolumn{2}{c}{--} & -- & 88.6$\pm$1.6 & 90.6$\pm$1.2  & 96.2$\pm$0.6 & 97.7$\pm$0.4 \\ 
\cmidrule(r){1-1} \cmidrule(l){2-8}
\multirow{3}{*}{3DMM$^*$\emph{+p.}} & \cmark  &  \xmark & 60.7$\pm$2.0 & 30.6$\pm$3.2 & 34.3$\pm$2.2 & 55.1$\pm$2.1 & 65.1$\pm$2.0 \\
& \xmark  & \cmark  & 71.1$\pm$1.8 & 39.5$\pm$4.8 & 49.8$\pm$2.5 & 69.5$\pm$1.4 & 76.8$\pm$1.0 \\
& \cmark  &  \cmark & 75.4$\pm$1.6 & 46.6$\pm$5.1 & 57.2$\pm$1.9 & 74.4$\pm$1.3 & 80.5$\pm$1.1\\
\cmidrule(r){1-1}
3DDFA\emph{+p.}& \cmark & \xmark & 43.3$\pm$2.5 & 12.5$\pm$1.9 & 16.7$\pm$1.9 & 38.3$\pm$2.7 & 51.3$\pm$3.0 \\
\cmidrule(r){1-1}
\multirow{3}{*}{Us \emph{+pool}}
& \cmark  &  \xmark & 86.0$\pm$1.7 & 55.9$\pm$5.5 & 72.3$\pm$1.4 & 88.0$\pm$1.4 & 91.8$\pm$1.1 \\
& \xmark  &  \cmark & 83.5$\pm$2.2 & 50.3$\pm$5.8 & 70.9$\pm$1.5 & 87.3$\pm$1.1 & 91.5$\pm$1.0 \\
& \cmark  &  \cmark & \tbf{87.0$\pm$1.5} & \tbf{60.0$\pm$5.6} & \tbf{76.2$\pm$1.8} & \tbf{89.7$\pm$1.0} & \tbf{92.9$\pm$1.0} \\
\bottomrule
\end{tabularx}\vspace{-2mm}
\caption{{\em IJB-A face verification and recognition.} Comparing our 3DMM regression with others, including baseline face recognition methods.$^*$ Denotes the same method used to produce 3DMM target values for our CNN training.}
\vspace{-4mm}
\label{tab:ijba}
\end{table}
\begin{figure*}[t]
  \includegraphics[width=.24\linewidth,clip,trim = 0mm 0mm 0mm 0mm]{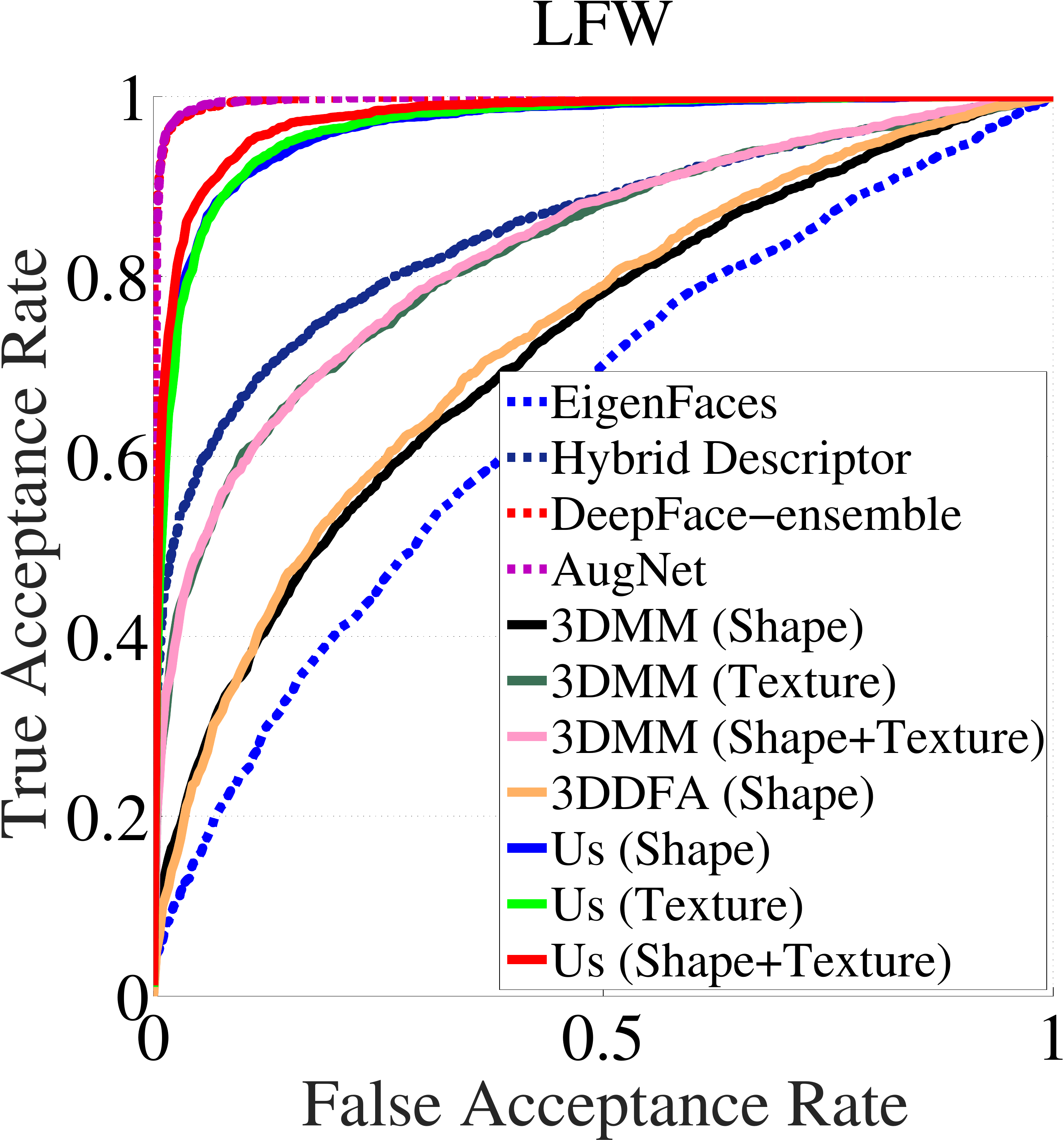}
   \includegraphics[width=.24\linewidth,clip,trim = 0mm 0mm 0mm 0mm]{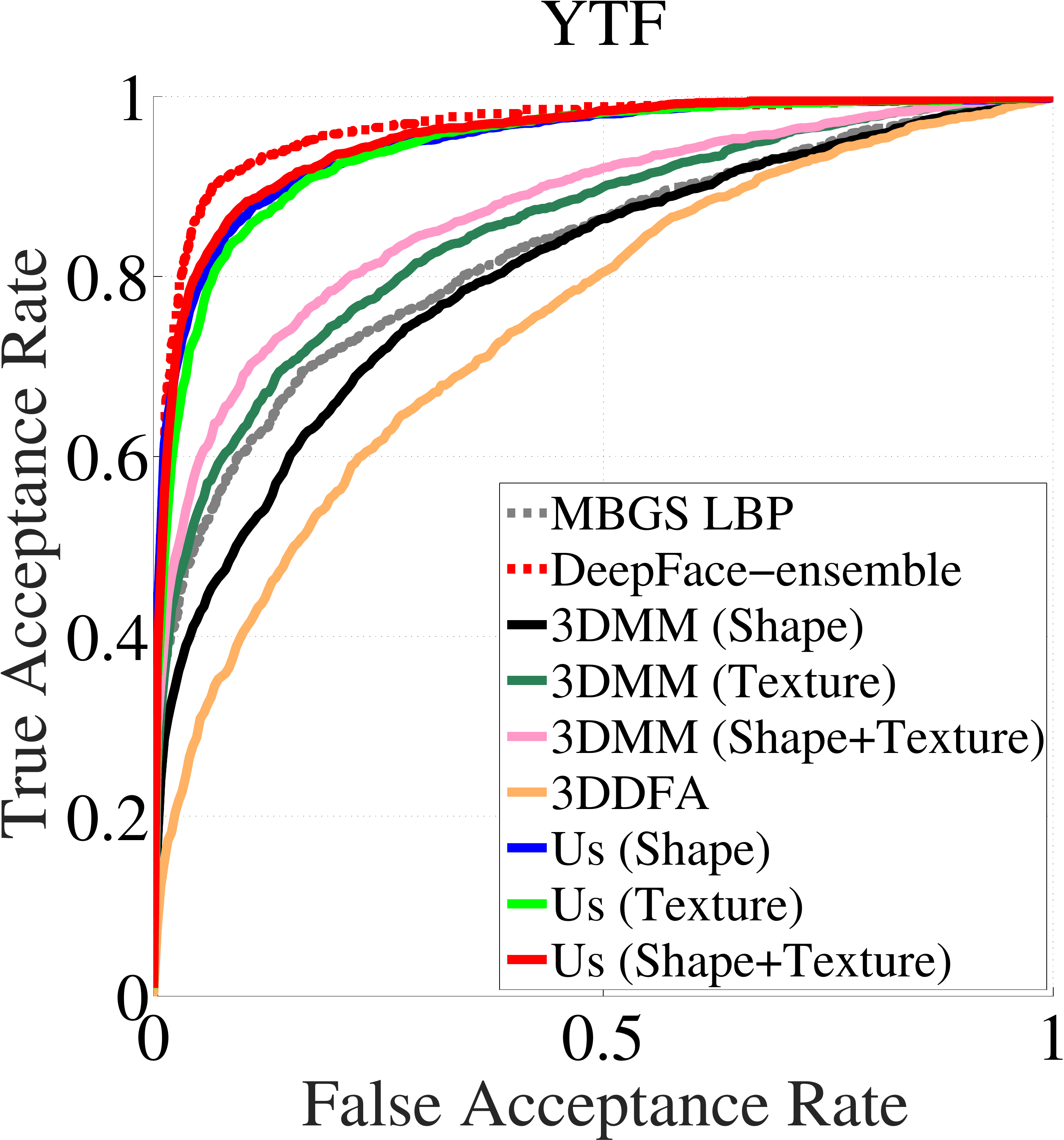}
  \includegraphics[width=.24\linewidth,clip,trim = 0mm 0mm 0mm 0mm]{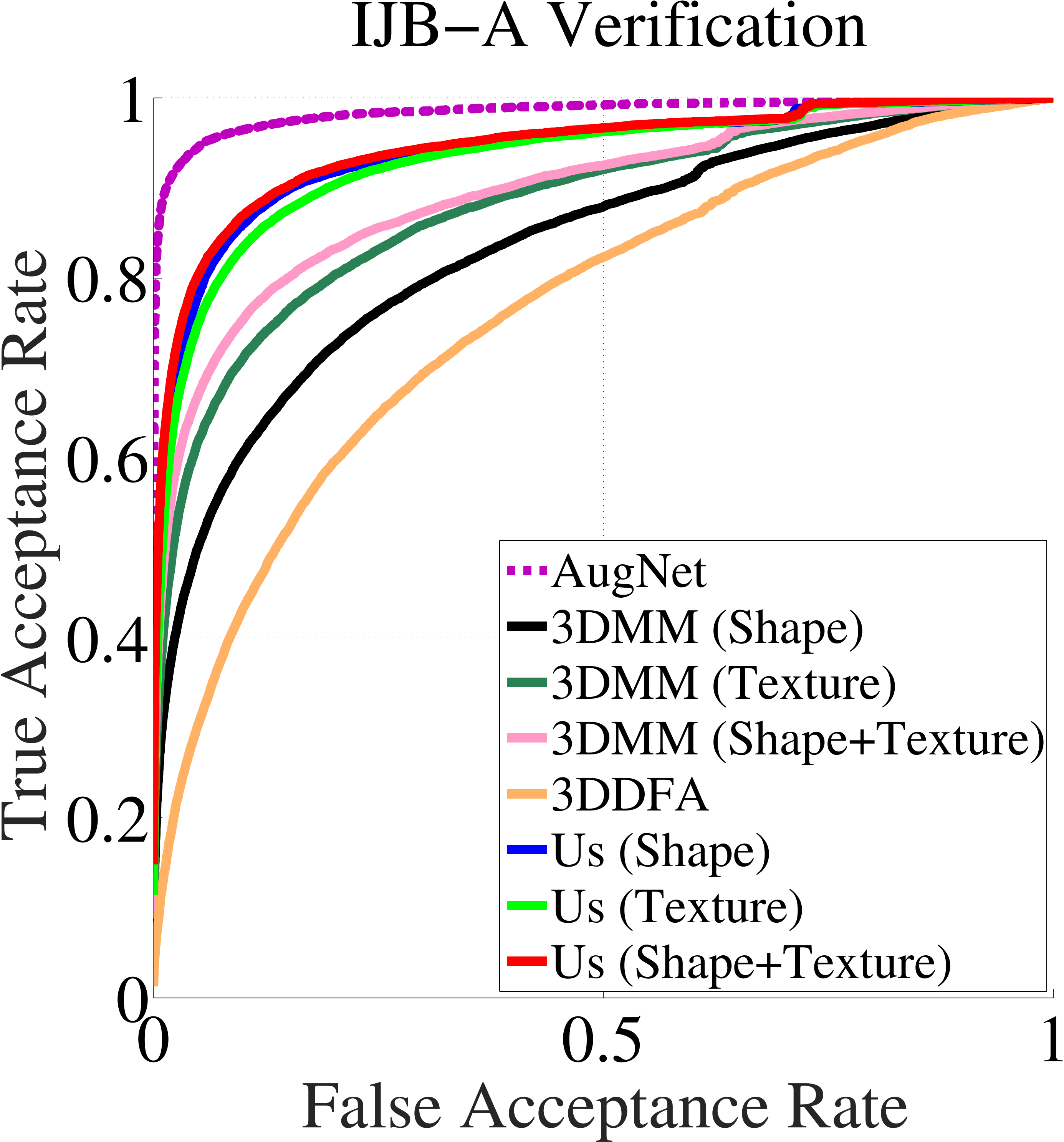}
   \includegraphics[width=.25\linewidth,clip,trim = 0mm 0mm 0mm 0mm]{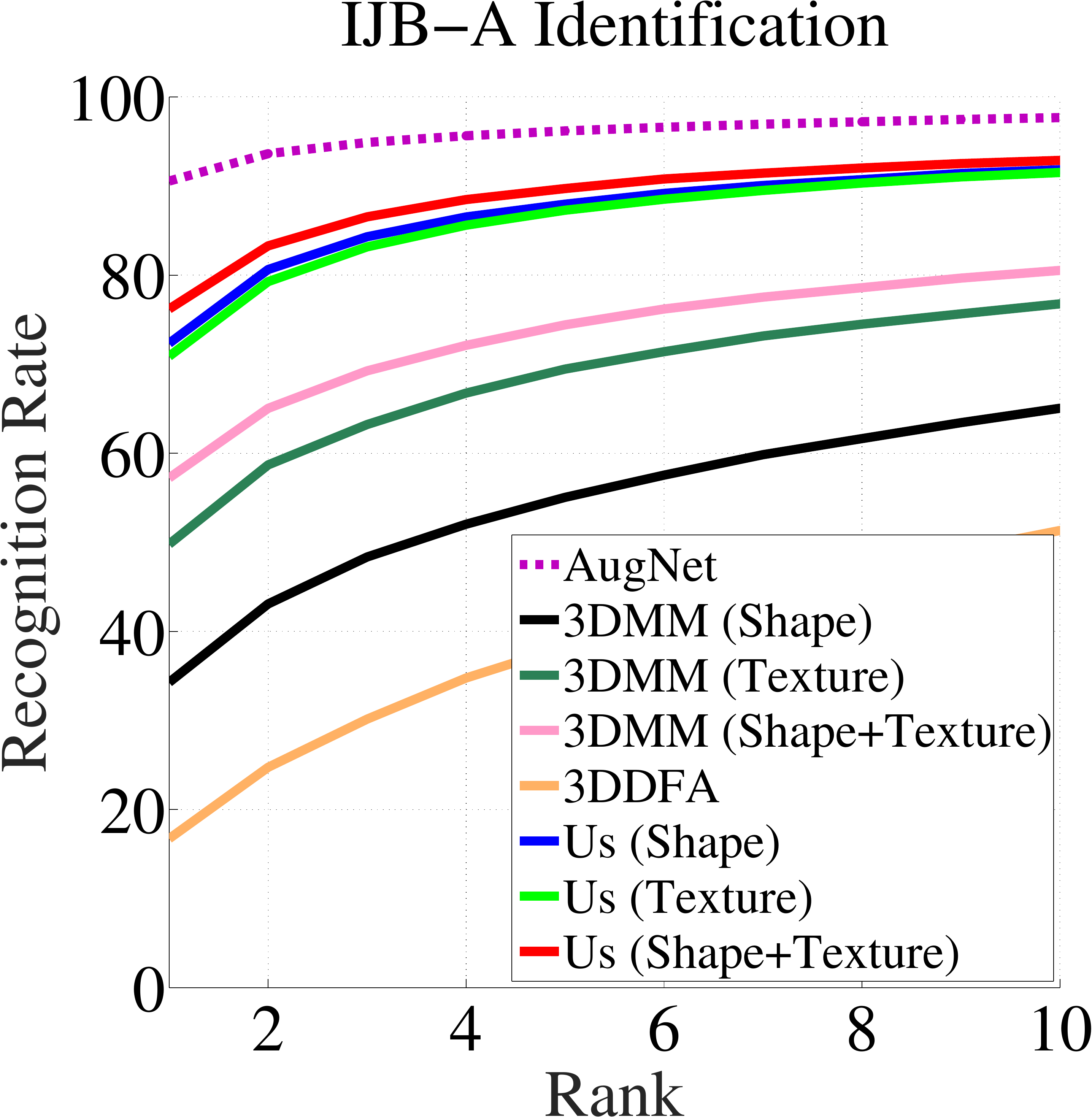}
   \caption{{\em Face verification and recognition results.} From left to right: Verification ROC curves for LFW, YTF, and IJB-A, and the recognition CMC for IJB-A.}
\label{fig:allcurves}
\vspace{-4mm}
\end{figure*}

These videos were used for single image and multi frame 3D reconstructions, comparing our method to existing alternatives. In these tests, estimated and ground truth shape parameters were converted to 3D using Eq.~(\ref{eq:3DMM}), cropped at a radius of 95$mm$ around the tip of the nose and globally aligned using the standard, rigid iterative closest point (ICP) method~\cite{besl1992method}, obtaining $X, X^*\subseteq \mathbb{R}^3$, respectively. They were additionally projected to a frontal view, obtaining depth maps $D_Q$ and $D_Q^*$. Estimation accuracy was then computed with standard error measures~\cite{hassner2013viewing,saxena08dist}:
\begin{itemize}
\item {\bf 3D Root Mean Square Error} (3DRMSE):\\$\sqrt{ \sum_i (X - X^*)^2} /N_v$
\item {\bf Root Mean Square Error} (RMSE):\\$\sqrt{ \sum_i (D_{Q_i} - D_{Q_i}^*)^2} /N_p$
\item $\mathbf{log_{10}}$: $|\log_{10}(D_Q)-\log_{10}(D_Q^*)|$
\item {\bf Relative error} (Rel): $|D_Q-D_Q^*|/|D_Q^*|$
\end{itemize}
Here, $N_v$ is the number of 3D vertices and $N_p$ the number of pixels in these representations.

Single view estimation was performed on the most frontal frame. Multi frame reconstructions were given the entire videos. Our multi frame results were produced by pooling 3DMM estimates from different frames, using Eq.~(\ref{eq:pooling}), with equal weights used for all frames. For all 3DMM fitting baselines~\cite{Bas:accvw16,huber:3dmm,romdhani2005estimating,Zhu2016Face}, we found that estimating shape, texture {\em and expression} parameters but using only shape and texture for comparisons, gave the best results. This approach was therefore used in all our tests.

Results are reported in Tab.~\ref{tab:micc}. Error rates are averaged across all videos and provided $\pm$ standard deviation. Our method is clearly the most accurate. Remarkably, both its single view and multiple frame versions outperform the method used to produce the training set target 3DMM labels (3DMM\emph{+pool}). This may be due to our use of such a large dataset to train the CNN and their known robustness to training label errors and noise~\cite{xie2016disturblabel}.

Our estimates are more accurate than the very recent state-of-the-art. This includes 3DDFA~\cite{Zhu2016Face} which fits 3DMM parameters by using a CNN to deal with large pose variations as well as~\cite{huber:3dmm} and~\cite{Bas:accvw16}. To better appreciate these numbers, note that our improvement over standard 3DMM fitting is comparable to their improvement over using a unmodified, generic Basel face shape~\cite{paysan09basel}.

\subsection{3DMM regression speed} Tab.~\ref{tab:micc} (rightmost column) also reports the average, per image runtime in seconds, required by the various methods to predict 3D face shapes. We compared our approach with iterative methods such as classic 3DMM implementations~\cite{Bas:accvw16,huber:3dmm,romdhani2005estimating}, the flow-based method of~\cite{hassner2013viewing} and also with a recent CNN based method~\cite{Zhu2016Face}.

As mentioned earlier, our method is render-free, without optimization loops which render the estimated parameters and compare them to the input photo. Unsurprisingly, at 0.088s ($\sim$11Hz), our CNN is {\em several orders of magnitude faster} predicting 3DMM parameters than most of the methods we tested. The second fastest method, by a wide gap, is the 3DDFA of~\cite{Zhu2016Face}, requiring 0.146s ($\sim$7Hz) for prediction.

Runtime was measured on two different systems. All our baselines required MS-Windows to run and were tested on an Intel Core i7-4820K CPU @ 3.7GHz with 16GB RAM and a NVIDIA GeForce GTX 770. Our method requires Linux and so was tested on an Intel Xeon CPU @ 3.60GHz, with 12 GB of RAM and GeForce GTX 590. Importantly, the system used to measure our runtime is the slower of the two. Our runtimes may therefore be exaggerated.

\subsection{Face recognition in the wild}\label{sec:exp:rec}
We next consider the robustness of our 3DMM estimates and how discriminative they are. We aim to see if our 3DMM estimates for different unconstrained photos of the same person are more similar to each other than to those of other subjects. An effective way of doing this is by testing our 3DMM estimates on face recognition benchmarks. We emphasize that our goal is {\em not} to set new face recognition records. Doing so would require competing with state of the art systems designed exclusively for that problem. We provide performances of relevant (though not necessarily state of the art) recognition systems only as a reference. Nevertheless, our results below are the highest we know of that were obtained with meaningful features (here, shape and texture parameters) rather than opaque representations.

Our tests use the pipeline described in Sec.~\ref{sec:matching} and report multiple recognition metrics for verification (in LFW and YTF) and identification metrics (in IJB-A). These metrics are verification accuracy, 100\%-EER (Equal Error Rate), Area Under the Curve (AUC), and recall (True Acceptance Rate) at two cut-off points of the False Alarm Rate (TAR-\{10\%,1\%\}). For identification we report the recognition rates at various ranks from the CMC (Cumulative Matching Characteristic). For each tested method we also indicate its use of estimated 3D shape and/or texture. Finally, bold values indicate best scoring 3D reconstruction methods.

\minisection{Labeled Faces in the Wild (LFW)}~\cite{LFWTech} results are provided in Tab.~\ref{tab:lfw-ytf} (top) and Fig.~\ref{fig:allcurves} (left). Evidently, the shapes estimated by 3DDFA~\cite{Zhu2016Face} are only slightly more robust and discriminative than the classical eigenfaces~\cite{turk:eigface}. Fitting 3DMMs using~\cite{romdhani2005estimating} does better, but falls behind the Hybrid method of~\cite{WHT:ECCVW08:DBMW}, one of the first results on LFW, now nearly a decade old. Both results suggest that the shapes estimated by these methods are unstable in unconstrained settings and/or are too generic. By comparison, recognition performances with our estimated 3DMM parameters is not far behind those recently reported by Facebook, using their multi-CNN approach trained on four million images~\cite{taigman2014deepface}.

\minisection{YouTube Faces (YTF)}~\cite{wolf:YTF} Accuracy on YTF videos is reported in Tab.~\ref{tab:lfw-ytf} (bottom) and Fig.~\ref{fig:allcurves} (mid-left). Though video frames in this set are often low in quality and resolution, our method performs well. It is outperformed by the Facebook CNN ensemble system~\cite{taigman2014deepface}, explicitly designed for face recognition, by an AUC gap of only $\sim$1\%. The 3DMM shapes and textures estimated by other methods perform far worst, with~\cite{romdhani2005estimating} doing only slightly better than the MBGS face recognition system~\cite{wolf:YTF}, which is the oldest result on that benchmark and~\cite{Zhu2016Face} falling far behind.

\minisection{IARPA Janus Benchmark A. (IJB-A)}~\cite{Klare_2015_CVPR}
~Released recently, IJB-A was designed to offer elevated challenges compared to other face recognition benchmarks. In particular, it presents faces in near profile poses, almost nonexistent in previous face sets. It further contains faces in extremely low resolution and often strongly affected by noise. 

We evaluated both the face verification (1:1) and recognition (1:N) protocols and report results in Tab.~\ref{tab:ijba} and Fig.~\ref{fig:allcurves} (mid-right, right). Here too, performances adopt the same pattern as in the previous two benchmarks, with 3D shapes estimated by 3DDFA~\cite{Zhu2016Face} performing far worst than other methods. Our own method performs quite well, though it is outperformed by a wide margin by the very recent face recognition system of~\cite{masi16dowe}, which was designed for this set.  

\begin{figure}[t]
\centering
\includegraphics[width=.98\columnwidth]{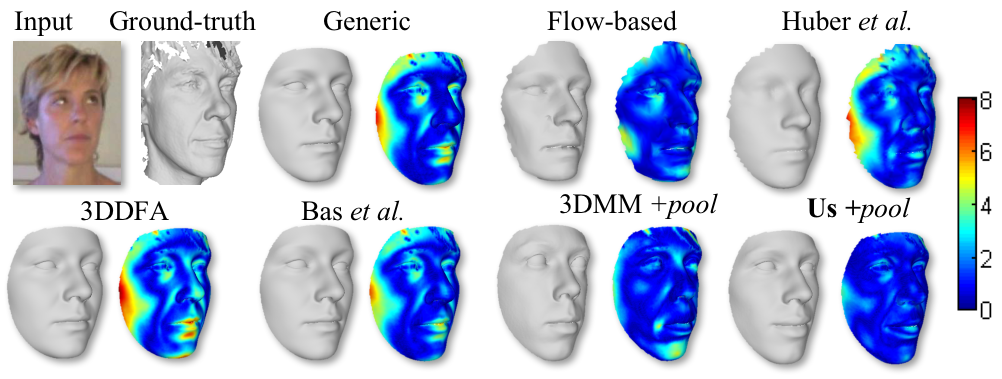}
\caption{{\em Qualitative comparison of surface errors,} visualized as heat maps with real world $mm$ errors on faces from MICC videos and their ground truth 3D shapes. Left to right, top to bottom: frame from input and 3D ground-truth shape; the generic face; estimates for flow-based method~\cite{hassner2013viewing}, Huber et al.~\cite{huber:3dmm}, 3DDFA~\cite{Zhu2016Face}, Bas et al.~\cite{Bas:accvw16}, 3DMM \emph{+pool}~\cite{romdhani2005estimating}, our method \emph{+pool}.
}\label{fig:heatmap}
\vspace{-4mm}
\end{figure}
\begin{figure}[t]
\centering
\includegraphics[width=.92\columnwidth]{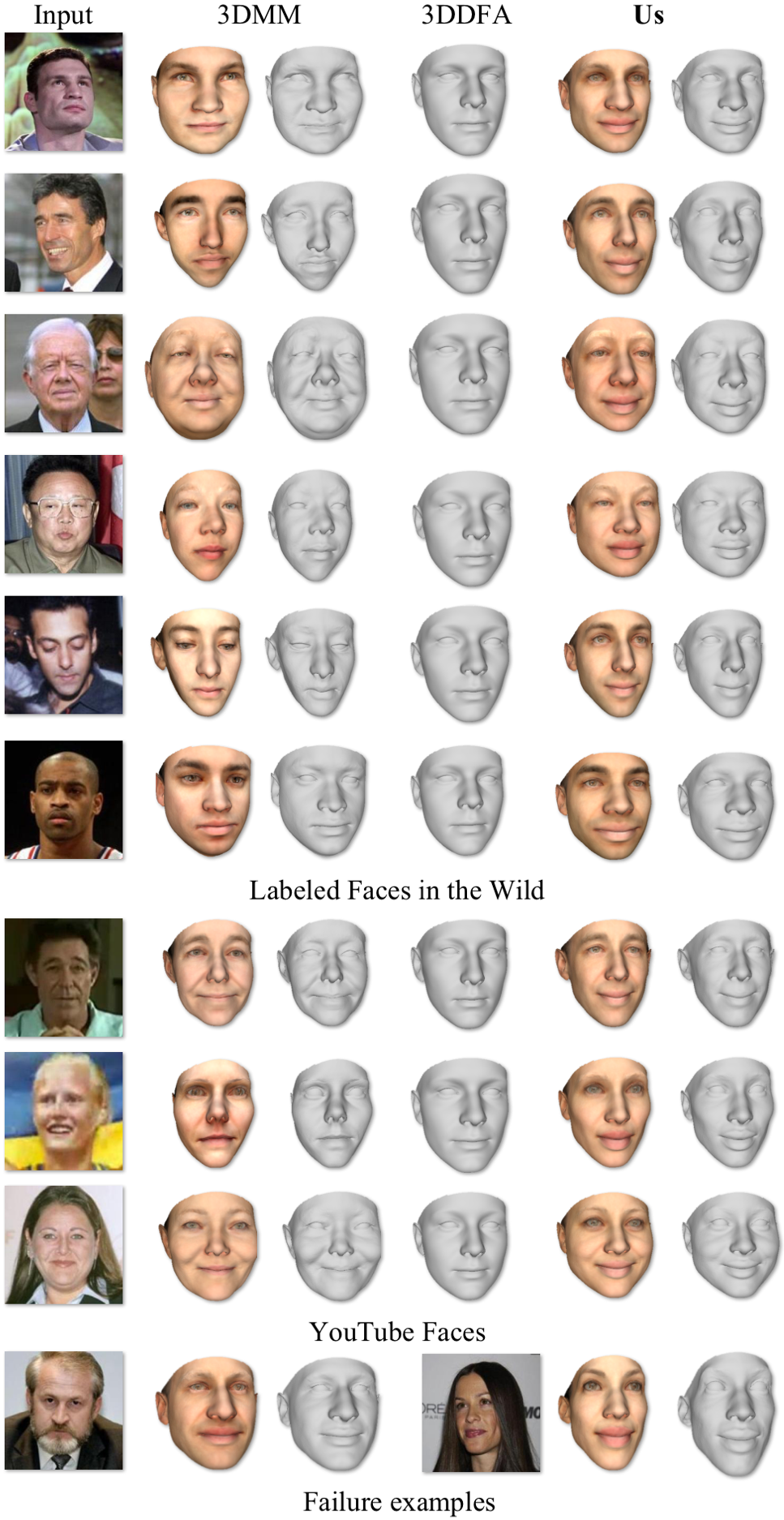}
\caption{
{\em Qualitative results}, produced by 3DMM~\cite{romdhani2005estimating}, 3DDFA~\cite{Zhu2016Face} and our method on still-images from LFW and single frames from YTF. Bottom: Two failure examples.}\label{fig:lfwquall}
\vspace{-5mm}
\end{figure}

\subsection{Qualitative Results} \label{sec:exp:qual}
Fig.~\ref{fig:heatmap} provides a qualitative comparison of the surface errors in $mm$ for different methods for a subject in the MICC dataset. Our method produces visually smaller errors compared to ground-truth. The areas around the nose and mouth in particular have very low errors, while other methods are more sensitive in these regions (e.g 3DDFA~\cite{Zhu2016Face}). We provide also qualitative 3D reconstructions of faces in the wild, using images from LFW and single frames from YTF videos. Fig.~\ref{fig:lfwquall} presents these results showing both rendered 3D shapes and (when available) also its estimated texture. These results show that our method generates more visually plausible 3D and texture estimates compared with those produced by other methods. Fig.~\ref{fig:heatmap} also shows a few failure cases, here due to facial hair which was missing from the original 3DMM representation and extreme out-of-plane rotation which produced a thin, unrealistic 3D shape.

\section{Conclusions}
We show that existing methods for estimating 3D face shapes may either be sensitive to changing viewing conditions, particularly in unconstrained settings, or too generic. Their estimated shapes therefore do not capture identity very well, despite the fact that true 3D face shapes are known to be highly discriminative. 

We propose instead to use a very deep CNN architecture to regress 3DMM parameters directly from input images. We provide a solution to the problem of obtaining sufficient labeled data to train this network. We show our regressed 3D shapes to be more accurate than those of alternative methods. We further run extensive face recognition tests showing these shapes to be robust to unconstrained viewing conditions and discriminative. Our results are furthermore the highest recognition results we know of, obtained with interpretable representations rather than opaque features. We leave it to future work to regress more 3DMM parameters (e.g., expressions) and design state of the art recognition systems using these shapes instead of the abstract features used by others.
 
\section*{Acknowledgments}
This research is based upon work supported in part by the Office of the Director of National Intelligence (ODNI), Intelligence Advanced Research Projects Activity (IARPA), via IARPA 2014-14071600011. The views and conclusions contained herein are those of the authors and should not be interpreted as necessarily representing the official policies or endorsements, either expressed or implied, of ODNI, IARPA, or the U.S. Government.  The U.S. Government is authorized to reproduce and distribute reprints for Governmental purpose notwithstanding any copyright annotation thereon.

{\small
\bibliographystyle{ieee}

}
 
\end{document}